\documentclass[nofootinbib,reprint,aps,amsmath,amssymb,floatfix,screen]{revtex4-1}
\usepackage{amsmath,amsfonts,bm}

\def\1{\bm{1}}

\DeclareMathAlphabet{\mathsfit}{\encodingdefault}{\sfdefault}{m}{sl}
\SetMathAlphabet{\mathsfit}{bold}{\encodingdefault}{\sfdefault}{bx}{n}

\DeclareMathOperator*{\argmin}{arg\,min}

\usepackage[colorlinks=true,linkcolor=blue,citecolor=blue,urlcolor=violet]{hyperref}
\usepackage{hhline}
\usepackage{multirow}
\usepackage{dsfont}       % simple URL typesetting
\usepackage{booktabs}       % professional-quality tables
\usepackage{nicefrac}       % compact symbols for 1/2, etc.
\usepackage{microtype}  
\usepackage{graphicx}
\usepackage[caption=false]{subfig}
\usepackage{textcomp}
\usepackage{tikz} 
\usepackage[algoruled,vlined]{algorithm2e}
\usepackage{setspace}

\usetikzlibrary{bayesnet}
\usetikzlibrary{calc}
\usetikzlibrary{arrows.meta} 

\begin{document}

\title{Task-Driven Data Verification via Gradient Descent}

% Authors must not appear in the submitted version. They should be hidden
% as long as the \iclrfinalcopy macro remains commented out below.
% Non-anonymous submissions will be rejected without review.

% The \author macro works with any number of authors. There are two commands
% used to separate the names and addresses of multiple authors: \And and \AND.
%
% Using \And between authors leaves it to \LaTeX{} to determine where to break
% the lines. Using \AND forces a linebreak at that point. So, if \LaTeX{}
% puts 3 of 4 authors names on the first line, and the last on the second
% line, try using \AND instead of \And before the third author name.

% \newcommand{\fix}{\marginpar{FIX}}
% \newcommand{\new}{\marginpar{NEW}}

\newcommand{\etal}{\emph{et al.}}

\newcommand{\alert}[1]{{\color{red} #1}}

\newcommand{\thr}[1]{{$\text{CDGD}_{#1}$}}
\newcommand{\fin}{$\text{CDGD}_{T\,}$}
\newcommand{\cl}{S_c}
\newcommand{\no}{S_n}

\newcommand{\sq}[1]{{$#1 \times #1$}}
% \acmPrice{15.00}
% \acmDOI{10.1145/1122445.1122456}
% \acmISBN{978-1-4503-9999-9/18/06}

\author{Siavash Golkar}
\email{golkar@nyu.edu}
\affiliation{New York University}

\author{Kyunghyun Cho }
\email{kyunghyun.cho@nyu.edu}
\affiliation{New York University \\
Facebook AI Research \\
CIFAR Azrieli Global Scholar}

\begin{abstract}
We introduce a novel algorithm for the detection of possible sample corruption such as mislabeled samples in a training dataset given a small clean validation set. We use a set of inclusion variables which determine whether or not any element of the noisy training set should be included in the training of a network. We compute these inclusion variables by optimizing the performance of the network on the clean validation set via ``gradient descent on gradient descent'' based learning. The inclusion variables as well as the network trained in such a way form the basis of our methods, which we call Corruption Detection via Gradient Descent (CDGD). This algorithm can be applied to any supervised machine learning task and is not limited to classification problems. We provide a quantitative comparison of these methods on synthetic and real world datasets. 
\end{abstract}

\maketitle

%%%%%%%%%%%%%%%%%%%%%%%%%%%%%%%%%%%%%%%%%%%%%%%%%%%%%%%%%%%%%%%%%%%%%%%%%%%%%%%%%%%%%%%%%%%%%%%%%%%%%%%%%%%%%%%%%%%%%%%%%%%%%%

\section{Introduction}

In recent years, we have experienced a rapid improvement in the performance of machine learning algorithms, brought about by a combination of increase in available computational resources and the availability of large amounts of data. Much research has been done on optimizing the performance of networks given the amount of available resources in terms of available computation power and data, however, ultimately, the generalization performance of any algorithm is bound by the quantity and quality of the dataset available. For the machine learning practitioner of today, the process of collecting and verifying the data remains the most tedious yet crucial part of the process. To this end, many solutions have been developed with the promise of providing higher quantities of data (e.g. Amazon Mechanical Turk), however these solutions often come at the cost of lower quality, with many mislabeled or otherwise erroneous data points mixed into the training set. It can therefore be argued that the act of verifying the validity of each individual sample, is the most time consuming part of the data analysis work-flow. As such, we believe any contribution in the direction of improving dataset reliability and speeding up sample verification would be of great practical utility to the machine learning community.

In this work, we tackle the problem of data verification by introducing a general method that can identify corrupt samples for any supervised learning task given a small clean validation set. The idea is that this would save much time for the practitioner who would then need to manually check the reliability of only a small portion of the samples, and then rely on an automated algorithm to find the erroneous samples in the remainder of the dataset. In simple terms, the idea behind this method is to use gradient descent to learn which subset of the noisy training data would lead to a network which performs best on the clean dataset. Any parts of the noisy examples which are hereby excluded, i.e. would lead to a poorer performance on the clean dataset, are identified as erroneous.

Our main contributions are:
\begin{enumerate}
    \item A general methodology for the detection of corrupt training samples given a small clean validation set, which can be applied to \emph{any} supervised machine learning task.
    \item A method to efficiently learn the coefficients of weighted samples via ``gradient descent on gradient descent'' based learning.
\end{enumerate}

The remainder of the paper is organized as follows: in Sec.~\ref{sec:motivation} we review a common work-flow scenario of data analysis projects and discuss where our data verification algorithm fits in. In Secs. \ref{sec:generalities} and \ref{sec:details} we provide the main idea and general considerations of our method. In Sec.~\ref{sec:practicals}, we discuss some of the practical considerations regarding various choices of hyperparameters and networks. We provide an empirical study of our method in Sec.~\ref{sec:experiments}, first on a synthetic dataset comprised of points on spheres in Sec.~\ref{sec:sphere} and then on the MNIST and CIFAR-10 \cite{CIFAR} benchmarks in Sec.~\ref{sec:mnist_cifar}.  We conclude in Sec.~\ref{sec:conclusion}. While we focus our empirical studies on classification problems for simplicity, our methodology is also applicable to other supervised machine learning tasks.

\subsection*{Related Work}

\paragraph{Outlier Removal} 

There exists an extensive literature on the problem of outlier detection and removal of corrupt training samples, ranging from the use of trimmed decision trees~\cite{outlierremoval_decisiontree}, training of filter algorithms~\cite{outlierremoval_filters}, estimation of noise distribution via a noise transformation~\cite{outlierremoval_noiselyaer}, the use of curriculum schemes with student/mentor networks~\cite{outlierremoval_curriculum}, and the use of non-trivial regularization~\cite{outlierremoval_air_reg}. More recently, methods have been proposed which utilize numerous techniques to approach this problem. These include: additional labels for label cleaning~\cite{outlierremoval_additional_labels}, a distillation framework to use side information label relations in knowledgegraphs~\cite{outlierremoval_distillation}, and  undirected graphical models representing the relationship between noisy and clean labels, trained in an semisupervised setting~\cite{outlierremoval_semisup2}. 

Most of these works, however, are  either unsupervised and struggle at distinguishing corruption from exceptions~\cite{outlierremoval_filters,outlierremoval_noiselyaer,outlierremoval_curriculum}, or are not applicable to the analysis of large datasets and neural networks~\cite{outlierremoval_decisiontree}. Furthermore, perhaps because of the proliferation of image classification benchmark papers, there has been an outsized attention to classification tasks and only to corruptions which are restricted to sample labels~\cite{outlierremoval_filters,outlierremoval_noiselyaer,outlierremoval_curriculum,outlierremoval_air_reg,outlierremoval_additional_labels,outlierremoval_distillation,outlierremoval_semisup2}. We thus identify the importance of a general purpose corruption detection algorithm that can be applied to any supervised learning task. This is the goal of this paper.

In all the aforementioned prior work regarding mislabeled data, the focus of attention has been predominantly on improving generalization performance on a small clean test set while neglecting the statistics of the detection of  the noisy data, even when the noisy/clean samples of the dataset in question are known. Indeed improved test accuracy and better detection of noisy samples do not necessarily go hand in hand, and scoring outlier detection methods solely based on test set performance runs the risk of confusing exceptions in the dataset (e.g. samples from the tail of a distribution) with corrupt data. This is especially true in cases where the test set is very small compared to the training set or the distribution is fat tailed. In this work we focus our analysis on the statistics of correctly identifying corrupt samples. We hope that this practice is picked up by other researchers in this field when possible.

\paragraph{Weighted samples} 

Initially proposed by~\cite{mackay1992information,bengio2000gradient}, gradient-based hyperparameter optimization has been further explored in~\cite{luketina2016scalable,maclaurin2015gradient}. More recently, it has been applied to various problems including~\cite{gbrain-weights} where it was used to improve generalization. We use sample weights in a manner similar to this work and~\cite{prior}, however, unlike these works which focus on the improvement of generalization on clean datasets we focus on the detection of the corrupt samples.

%%%%%%%%%%%%%%%%%%%%%%%%%%%%%%%%%%%%%%%%%%%%%%%%%%%%%%%%%%%%%%%%%%%%%%%%%%%%%%%%%%%%%%%%%%%%%%%%%%%%%%%%%%%%%%%%%%%%%%%%%%%%%%
\section{Motivation}
\label{sec:motivation}
%%%%%%%%%%%%%%%%%%%%%%%%%%%%%%%%%%%%%%%%%%%%%%%%%%%%%%%%%%%%%%%%%%%%%%%%%%%%%%%%%%%%%%%%%%%%%%%%%%%%%%%%%%%%%%%%%%%%%%%%%%%%%%
In this section we review a  work-flow scenario which is common to many data analysis projects and discuss how our methodology for improving the quality of dataset fits in with this overall scheme. Generally speaking, the main steps of machine learning projects fall under four main categories. The first is data collection and annotation. This step can be the most time consuming part of the process and is often outsourced to third party commercial sources such as Amazon Mechanical Turk. This is followed by a conjecture of a model which can fit the data, either by a careful analysis or often by educated guess work. This step is greatly simplified by the availability of powerful software packages and online tools. In the third step, the model is fitted to the data using stochastic gradient descent or other learning algorithms. If the model fits the data as expected, we finally move on to to the final step of analyzing and extracting information from the data and the fitted model. If, however, the model fails to fit the data properly, or otherwise fails to satisfy some consistency conditions, e.g. does not generalize well to unseen data or does not satisfy other perhaps theoretical constraints, we go back to the drawing board and come up with a modified model in the second step and reiterate the following steps.

The process described above would be the work-flow of data analysis projects in an idealized environment, where the data collected in the first step is assumed to be correctly annotated and otherwise error free. In actual practice, much of the time of machine learning practitioners in real world applications is spent on the verification of data collected by outsourced parties (Fig.~\ref{fig:man-verify}). An alternative to this approach, which we will pursue in this paper, is for the data scientist to verify the validity of only a small portion of the data and use this smaller clean dataset as a means to detect the erroneous samples of the rest of the collected data. We can then take the output of this method to more efficiently guide the manual verification of the data if we so choose (Fig.~\ref{fig:ML-verify}). 

	\begin{figure*}[ht]
		\centering
		
		\hfill
		\begin{minipage}{0.45\textwidth}
		\centering
		\vspace{14pt}
    		\begin{tikzpicture}
    		\node[det, shape=rectangle, scale=0.8] (DC) {\;Data Collection \;};
    		\node[det, shape=rectangle, scale=0.8, below=0.5cm of DC] (verify1) {\;Manual Verification\;};
    		\node[det, shape=rectangle, scale=0.8, right=0.5cm of DC] (model) {\;Model Building\;};
    		\node[det, shape=rectangle, scale=0.8, below=0.5cm of verify1, xshift=2cm] (train) {\;Training \& Analysis\;};
    		
    		\path[-Stealth]
    		(DC) edge (verify1)
    		(verify1) edge (train)
    		(model) edge (train);
    		
            \draw [-Stealth] (train.east)  .. controls (3.5,-2.1)  and (4.3,-0.4) .. (model.east);
    		
    		\end{tikzpicture}
    		\vspace{14pt}
    		\caption{Data analysis work-flow.}
    		\label{fig:man-verify}
		\end{minipage}
		\begin{minipage}{0.45\textwidth} 
		\centering
    		\begin{tikzpicture}
    		\node[det, shape=rectangle, scale=0.8] (DC) {\;Data Collection \;};
    		\node[det, shape=rectangle, scale=0.8, below=0.5cm of DC] (verify1) {\;Manual Verification\;};
    		\node[det, shape=rectangle, scale=0.8, right=0.5cm of DC] (model) {\;Model Building\;};
    		\node[det, shape=rectangle, scale=0.8, below=0.5cm of verify1, xshift=2cm] (verify2) {\;ML Verification\;};
    		\node[det, shape=rectangle, scale=0.8, below=0.5cm of verify2] (train) {\;Training \& Analysis\;};
    		
    		\path[-Stealth]
    		(DC) edge (verify1)
    		(verify1) edge (verify2)
    		(model) edge (verify2)
    		(verify2) edge (train);
    		
            \draw [-Stealth] (verify2.west)  .. controls (-1,-1.7)  and (-2,-1.3) .. (verify1.west);
            \draw [-Stealth] (train.east)  .. controls (3.5,-2.1)  and (4.3,-0.4) .. (model.east);
    		
    		\end{tikzpicture}\caption{Work-flow with task-based sample verification.}
    		\label{fig:ML-verify}
		\end{minipage}
		\hfill
		
		%\label{fig:workflow}\caption{A comparison of data analysis workflows with and without ML based data verification.}
	\end{figure*}

The methods generally used for outlier detection and removal of examples with corrupt data generally fall under two main categories. First, we can treat this problem as an unsupervised anomaly detection task to try and find outliers in the dataset~\cite{outlierremoval_filters,outlierremoval_noiselyaer,outlierremoval_curriculum}. However, if we assume the existence of a smaller uncorrupted dataset, we can  treat this problem as supervised training using the model conjectured in the model building step. An example of one such approach would be to fit the model to the small clean dataset, and use the resulting trained network to analyze the noisy outsourced data. Alternatively we can use the model to fit to the noisy dataset, using the clean validation set as a validation set for early stopping and hyperparameter selection. These two methods are the verification tools commonly used by practitioners while searching for erroneous data points and not surprisingly outperform unsupervised methods which do not need a validation set comprised of uncorrupted samples.

In this paper we introduce a new  method in the second category of supervised learning. We propose to use gradient descent to determine which subset of the noisy dataset would lead to the best performance of the model on the verified clean set. That is, for every sample $x_i$ in the noisy dataset we learn a coefficient $\alpha_i\in[0,1]$ whose value determines whether or not it should be included when training the network which performs optimally on the clean set. In essence, we use the loss function evaluated on the noisy set to determine the parameters of the model, and then use the loss function on the validation set to determine the value of the coefficients~$\alpha_i$. 

In what follows we demonstrate our methodology on classification experiments. However, as described in Sec.~\ref{sec:generalities}, our method can be equally applied to other supervised machine learning problems.

%%%%%%%%%%%%%%%%%%%%%%%%%%%%%%%%%%%%%%%%%%%%%%%%%%%%%%%%%%%%%%%%%%%%%%%%%%%%%%%%%%%%%%%%%%%%%%%%%%%%%%%%%%%%%%%%%%%%%%%%%%%%%%
\section{Methodology Generalities} \label{sec:generalities}
%%%%%%%%%%%%%%%%%%%%%%%%%%%%%%%%%%%%%%%%%%%%%%%%%%%%%%%%%%%%%%%%%%%%%%%%%%%%%%%%%%%%%%%%%%%%%%%%%%%%%%%%%%%%%%%%%%%%%%%%%%%%%%

Let us assume that we are given two datasets: first, a large noisy dataset $\no$, comprised of samples that may or may not have been mislabeled, and second, a smaller clean dataset $\cl$, whose elements have been verified to be error-free. We would like to use the small clean dataset $\cl$  to find the mislabeled or erroneous elements of the large noisy dataset $\no$. 

First, consider training a network on the noisy dataset $\no$ and look at its performance on the clean dataset $\cl$. In other words we derive the optimal parameters of the network $\hat\theta$ by optimizing the loss function on $\no$, then evaluate the loss with these optimal parameters on $\cl$:
\begin{align*}
    L_c = L(\hat{\theta}; \cl),
\end{align*}
where $L_c$ denotes the loss computed on the clean dataset $\cl$, and 
\begin{align*}
    \hat{\theta} = \arg\min_{\theta} L(\theta; \no).
\end{align*}
Clearly, if the dataset $\no$ is very noisy, $L_c$, the loss function on the clean set, would not be close to its optimal value. However, if we had a way to slowly purge more and more erroneous samples from $\no$, and derive the network parameters $\hat\theta$ on this less noisy dataset, $L_c$ will get closer and closer to the optimial performance on $L_c$. We can therefore rephrase the problem of cleaning up $\no$ as finding the subset $M\subset\no$ which would give the lowest loss on the validation set $\cl$. That is, we perform the above procedure on all subsets $M$: 
\begin{align*}
    L_c(M) = L(\hat{\theta}(M); \cl),
\end{align*}
where
\begin{align*}
        \hat{\theta}(M) = \arg\min_{\theta} L(\theta; M),
\end{align*}
and take the subset $M^*$ which affords the lowest loss $\hat{\theta}(M)$ as the cleaned up version of $\no$.  The compliment set $\no\backslash M^*$ would therefore be our best guess at the noisy elements of $\no$. This approach is not practically feasible, however,  as the number of the possible subsets $M$ that we would need to explore is exponentially large in the size of $\no$. 

To proceed, let us define, for each element of the noisy dataset $x_i \in $ $\no$, a discrete variable $\alpha_i \in \{0,1\}$ which we call the inclusion variable. We can rewrite the loss on the subset $L(\theta,M)$ as:
\begin{align*}
    L(\theta,M) = L^{\alpha}(\theta; \no) = \frac{1}{N} \sum_{n=1}^N \alpha_n\, l(x_n; \theta),
\end{align*}
with 
\begin{align*}
    \alpha_n = \begin{cases}
    1 & x_n \in M \\
    0 &x_n \notin M
    \end{cases}\;,
\end{align*}
where $N$ is the number of elements in $\no$ and $l(x_n;\theta)$ is a per-sample loss. With this definition, we can further rewrite the optimally clean subset $M^*$ as:
\begin{align}
\label{eq:original_opt}
M^* = \{x_n\,|\,\alpha^*_n = 1 \},
\end{align}
where
\begin{align*}
\alpha^* = \argmin_{ \alpha \in \left\{ 0, 1\right\}^N } L(\arg\min_{\theta} L^\alpha(\theta; \no); \cl).
\end{align*}
This rewriting, however, does not make the underlying problem any simpler. We thus propose, in this paper, to relax the inclusion variable $\alpha \in \left\{ 0, 1\right\}^N$ to be continuous, i.e. $\alpha \in \left[ 0, 1\right]^N$. The optimal \emph{soft} inclusion parameters $\alpha^*$ can be derived by:
\begin{align}
\label{eq:proposed_opt}
\alpha^* = \argmin_{ \alpha \in [0, 1]^N } L(\arg\min_{\theta} L^\alpha(\theta; \no); \cl).
\end{align}
Since all the variables of this optimization problem are differentiable, we can use gradient descent to solve it.

The inclusion variables $\alpha^*$ are no longer integer valued. In order to read off the noisy/clean samples from $\alpha^*$, we propose two different methods. The first method, which we call Corruption Detection via Gradient Descent based on inclusion parameters $\alpha$ or \thr{\alpha} for short, considers all samples with $\alpha^*_i < 0.5$ as noisy.\footnote{
    In App.~\ref{app:pnr} we examine the behavior of this method as we change the threshold parameter away from 0.5.
} 
In the second method, we train a network using the real valued inclusion variables $\alpha^*$ and determine the noisy samples to be the elements of $\no$ which are classified incorrectly with this classifier. We call this method \fin (T for network Trained with the final inclusion parameters $\alpha^*$.)

We compare the performance of these two methods to two baseline algorithms often used by practitioners.  The first is to simply train a network on the entire $\no$ set, using the clean dataset $\cl$  for early stopping.\footnote{
    This method generally performs better than training on the combined noisy and clean datasets $\no \cup \cl$ since we forego the possibility of early stopping when we include $\cl$ in the training set.
}
Second, we train a network solely on the clean set $\cl$. In both methods we consider the noisy data to be those which the respective network does not classify correctly.  We call these two methods the $\no$ baseline and $\cl$ baseline respectively in reference to which dataset they are trained on. 

In non-classification problems, \fin as well as $\no$ and $\cl$ baselines can be generalized in a manner appropriate to the task. For example, for translation tasks, noisy samples can be taken as those which are not translated correctly or in regression tasks we can define noisy samples as those with loss larger than some threshold.

%%%%%%%%%%%%%%%%%%%%%%%%%%%%%%%%%%%%%%%%%%%%%%%%%%%%%%%%%%%%%%%%%%%%%%%%%%%%%%%%%%%%
\section{Sample Verification by Gradient Descent by Gradient Descent} 
\label{sec:details}
%%%%%%%%%%%%%%%%%%%%%%%%%%%%%%%%%%%%%%%%%%%%%%%%%%%%%%%%%%%%%%%%%%%%%%%%%%%%%%%%%%%%

The proposed optimization problem in Eq.~\eqref{eq:proposed_opt} can be split into two stages:
\begin{align*}
    \underbrace{\argmin_{ \alpha \in \left[ 0, 1\right]^N } L(
    \underbrace{\arg\min_{\theta} L^\alpha(\theta; D)}_{\text{inner optimization}}
    ;
    D_{\text{val}})}_{
    \text{outer optimization}}.
\end{align*}
Because this is now a continuous optimization problem, and the objective functions in both inner and outer stages are differentiable, we can use gradient descent to solve this problem with respect to the soft inclusion variables $\alpha$. We explain this gradient-descent-based approach in more detail below.

\paragraph{Inner Optimization}

The inner optimization procedure performs stochastic gradient descent (SGD) on $\theta$, generating a chain of $\theta$'s that are functions of $\alpha$. That is,
\begin{align}
\label{eq:inner-sgd}
    \theta^i = \theta^{i-1} - \eta \nabla_{\theta} L^{\tilde{\alpha}}(\theta^{i-1}; \tilde{D}),
\end{align}
where $\tilde{D} \subset \no$ is a minibatch, and $\tilde{\alpha}$ is a subset of $\alpha$ associated with the training examples in $\tilde{D}$. Eventually at the end of SGD, we end up with the final $\hat{\theta}=\theta^I(\alpha)$, where $I$ is the number of SGD steps. Looking at Eq.~\eqref{eq:inner-sgd} carefully, it becomes clear that this is nothing but a recurrent network with $\theta$ as a hidden state, $\tilde{D}$ as an input and $\tilde{\alpha}$ as a set of parameters, and that we can immediately use backpropagation-through-time~\citep{werbos1990backpropagation} to compute a series of the Jacobians of $\theta^i$ with respect to all the influence variables $\alpha$, for which we use 
\[
\left( \frac{\partial \theta^1}{\partial \alpha}\,,\, \cdots \,,\, \frac{\partial \theta^I}{\partial \alpha} \right)
\]
as a shorthand.

% Since all the gradient descent steps in Eq.~\eqref{eq:inner-sgd} are differentiable, we can compute
% \begin{align*}
%     \nabla_{\alpha} \hat{\theta} 
%     = \sum_{i=1}^{I} \nabla_{\alpha} \theta^i,
% \end{align*}
% where $I$ is the number of SGD steps taken. 

\paragraph{Outer Optimization}

Although the proposed optimization problem in Eq.~\eqref{eq:proposed_opt} only involves minimizing the loss on the clean dataset $L(\theta^I; \cl)$ with respect to the final parameter $\theta^I$, we are interested also in the influence of the inclusion parameters on each and every step of SGD. This is due to the common practice of early stopping and using a fixed, often adaptive, number of SGD steps. That is, we want to estimate the influence of each training example regardless of how many SGD steps would be taken to train a network. We achieve this by writing the outer optimization problem as
\begin{align*}
    \min_{\alpha \in [0,1]^N} \sum_{i=1}^I
    L(\theta^i; \cl),
\end{align*}
given $(\theta^1, \ldots, \theta^I)$ from the inner optimization above. 

We then take a gradient descent step to solve the above optimization problem, starting from $\alpha$ from the previous outer optimization. The gradient is the sum of per-iteration gradients, each of which is computed as\footnote{
Instead of the gradient and Jacobian, we use the partial derivative notations to make the equation less cluttered and clearer. 
}
\begin{equation*}
    \frac{\partial L(\theta^i; \cl)}
    {\partial \alpha}
    = \frac{\partial \theta^i}
    {\partial \alpha}
    \frac{\partial L}
    {\partial \theta^i},
\end{equation*}
where the first term can be efficiently computed via backpropagation-through-time, as discussed earlier. This computation requires calculating the gradient of various gradients in the inner optimization process, which is equivalent to computing the Hessian-vector product, which is done efficiently by R-Op~\citep{pearlmutter1994fast}.

\paragraph{Overall Procedure}

The proposed approach performs $N_{\text{out}}$ iterations of the outer optimization procedure. Within each outer iteration, we run the inner optimization procedure with $N_{\text{in}}$ steps of SGD. We generally use separate schemes for setting the step size $\eta$ and optimization methods in each of inner and outer loops, as long as the adaptive learning rate scheme in the inner optimization procedure maintains the differentiability of the inner SGD steps. The algorithm is summarized in Alg.~\ref{alg:verify}, where we have omitted minibatching inside the inner loop for simplicity.

\begin{algorithm}[t]
\SetKwFor{RepTimes}{repeat}{times}{end}
\SetKwInOut{Input}{input}\SetKwInOut{Output}{output}

\BlankLine
\Input{Noisy dataset $\no$, clean dataset $\cl$}
\Output{Soft inclusion parameters $\alpha$}

\BlankLine
$\alpha \leftarrow 1$\;
\RepTimes{$N_{\text{out}}$}{
    $\theta \leftarrow$ random initialization\;
    $L_\text{sum} \leftarrow 0$\;
    % \setstretch{1.1}
    \RepTimes{$N_{\text{in}}$}{
        \setstretch{1.3}
        $L^{\alpha}(\theta; \no) \leftarrow \frac{1}{N} \sum_{n=1}^N \alpha_n\, l(x_n; \theta)$\;
        \setstretch{1.2}
        $\theta \leftarrow \theta - \eta_{\text{in}} \nabla_{\theta} L^{\alpha}(\theta; \no)$\;
        $L_\text{sum} \leftarrow L_\text{sum}+ L(\theta; \cl)$\;
        \setstretch{1.1}
    }
    \setstretch{1}
    $\alpha \leftarrow \alpha - \eta_{\text{out}} \nabla_{\alpha} L_\text{sum}$\;
    
    \For{$\alpha_i \in \alpha$}{
        \bf{if} $\alpha_i<0$ \bf{:} $\alpha_i\leftarrow 0$\;
        \bf{if} $\alpha_i>1$ \bf{:} $\alpha_i\leftarrow 1$\;
    }
}
\caption{Data verification}
\label{alg:verify}
\end{algorithm}

\section{Practical Considerations}
\label{sec:practicals}

There are a few optimization decisions that must be made in advance, such as how the neural network would be initialized for the inner-optimization process, how training examples must be shuffled and which optimizer to be used. Furthermore, the computational complexity of the proposed approach is clearly high considering the inner optimization procedure of the proposed approach needs to keep track of the $\alpha$ dependence during the training a full neural network from scratch until a certain convergence criterion is met. We discuss here some of the decisions we make, and also describe how we alleviate the issue of computational complexity.

As described in the previous section, each step of the outer optimization loop starts with an untrained network and takes $N_{in}$ steps of the inner optimization loop. At this stage we are presented with a number of choices regarding the initialization of the network, minibatch sizes, whether or not to shuffle the training set for different inner loop epochs and others. Generally, these choices are made based on maximizing the performance of the trained network in the inner loop. Here we discuss the effect of these choices on the performance of the outer loop. 

\paragraph{Parameter initialization}

We find that it is important to randomize the initial parameters $\theta^1$ of the network at the beginning of each inner loop optimization. The alternative, that is to start with the same initial parameters after each outer loop step, can lead to over-fitting the inclusion parameters $\alpha$ to a particular initialization. For example, in a classification problem, if the initial parameters are such that they favor a specific class, the $\alpha$ can converge to values which assign higher weights to examples of other classes in order to compensate for this initial bias.

\paragraph{Batch sizes and shuffling}

Shuffling the training examples and using smaller batch sizes are known to increase the generalization performance by increasing stochasticity during training~\cite{minibatch,lecun1998efficient}. With regards to our algorithm, one might worry that using small batch sizes in the inner loop has the possible disadvantage of treating different training samples unequally, e.g. examples seen earlier might be seen as more or less important than ones seen later in an epoch. In practice, however, we find that in our examples, using smaller batch sizes leads to faster overall training times and shuffling leads to a small improvement in performance.

\paragraph{Truncated Backpropagation}

Keeping track of the full $\alpha$ dependence of the network parameters $\theta^i$ is both computationally expensive and memory hungry. In fact, in our current implementation in PyTorch,\footnote{
\url{https://pytorch.org/}
} 
the memory requirement and computation requirementstime of this algorithm grow respectively linearly and quadratically in the number of inner loop steps. This makes dealing with large datasets and models difficult. 

\begin{algorithm}[t]
\SetKwFor{RepTimes}{repeat}{times}{end}
\SetKwInOut{Input}{input}\SetKwInOut{Output}{output}

\BlankLine
\Input{Noisy dataset $\no$, clean dataset $\cl$}
\Output{Soft inclusion parameters $\alpha$}

\BlankLine
$\alpha \leftarrow 1$\;
\RepTimes{$N_{\text{out}}$}{
    $\theta \leftarrow$ random initialization\;
    $\delta\alpha \leftarrow 0$\;
    % \setstretch{1.1}
    \RepTimes{$N_{\text{in}}$}{
        \setstretch{1.3}
        $L^{\alpha}(\theta; \no) \leftarrow \frac{1}{N} \sum_{n=1}^N \alpha_n\, l(x_n; \theta)$\;
        \setstretch{1.2}
        $\theta \leftarrow \theta - \eta_{\text{in}} \nabla_{\theta} L^{\alpha}(\theta; \no)$\;
        $\delta \alpha \leftarrow \delta\alpha + \nabla_{\alpha} L(\theta; \cl)$\;
        \setstretch{1.1}
    }
    \setstretch{1}
    $\alpha \leftarrow \alpha - \eta_{\text{out}} \delta\alpha$\;
    
    \For{$\alpha_i \in \alpha$}{
        \bf{if} $\alpha_i<0$ \bf{:} $\alpha_i\leftarrow 0$\;
        \bf{if} $\alpha_i>1$ \bf{:} $\alpha_i\leftarrow 1$\;
    }
}
\caption{Data verification with gradient truncation}\label{alg:truncated}
\end{algorithm}

In order to deal with this problem, we implement a checkpointing scheme, analogous to truncated backpropagation through time, where we regularly truncate the dependence of the parameters $\theta^i$ on $\alpha$. In practice, at fixed epochs during the inner loop training, we compute the gradient of the validation loss up to this point with respect to $\alpha$. Starting from the next epoch, treat the inner loop parameter $\theta^n$ as independent of both $\alpha$ and $\theta^i$ with $i<n$. Then, at the end of the inner loop training, we sum these gradients and perform the $\alpha$ update all at once (Alg.~\ref{alg:truncated}). This process has the benefit of greatly speeding up training and capping the memory consumption of inner-loop training. However, this truncation comes at the cost of losing the full dependence of $\theta$'s on the inclusion variables $\alpha$. In cases where it is possible to carry out our method without truncation, we see that implementing the truncation leads to longer training times and often worse final performance of the method. However, in many real world scenarios, the application of our method without truncation would simply not be possible. 

\section{Experiments}
\label{sec:experiments}

In what follows we 
provide an empirical exploration of our methodology. We  focus on three statistics regarding the correct detection of corrupt samples: precision, recall and F1 score. Low recall scores indicate the tendency for models to memorize corrupt data, and low precision scores indicate a method which designates many clean samples as corrupt. A good methodology would have both high precision and high recall scores which would lead to a high F1 score (geometric mean of precision and recall). We therefore use the F1 score as the deciding factor when comparing different methods.

\begin{figure}[t]
    % \centering
    % \fbox{
    \includegraphics[clip, trim=2.8cm 3.8cm 2.1cm 3.5cm,width=0.4\textwidth]{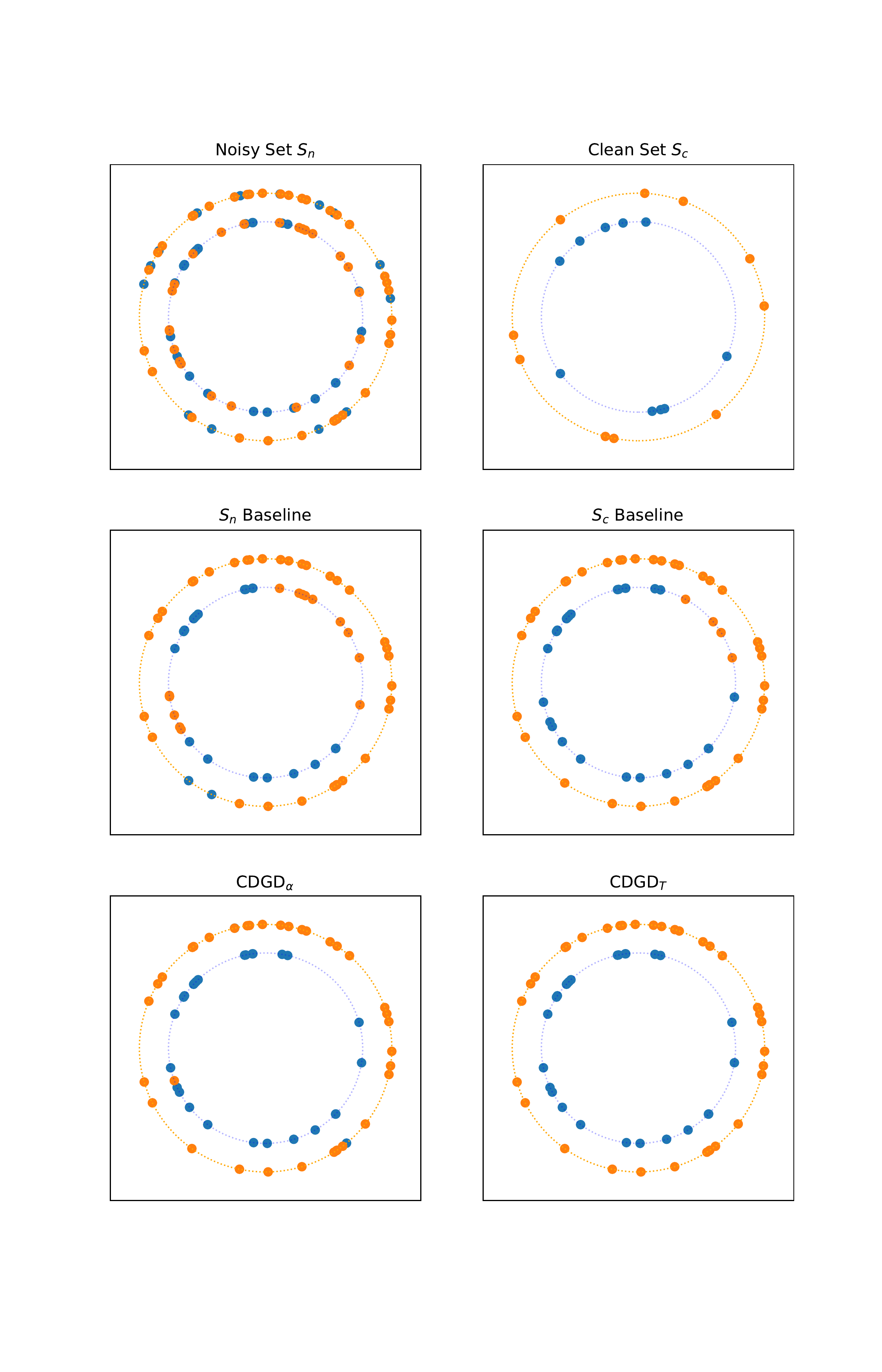} 
    % }
    \caption{Typical sample data and results for a circle (d=2). Blue and orange dots respectively denote samples with labels 0 and 1. Blue dots on the outer circle and orange dots on the inner circle are mislabeled samples. The noisy and clean datasets $\no$ and $\cl$ are given at the top. The rest of the plots denote the results of cleaning up the $\no$ dataset using the denoted algorithm.}
    \label{fig:sphere_demo}
\end{figure}

%%%%%%%%%%%%%%%%%%%%%%%%%%%%%%%%%%%%%%%%%%%%%%%%%%%%%%%%%%%%%%%%%%%%%%%%%%%%%%%%%%%%%%%%%%%%%%%%%%%%%%%%%%%%%%%%%%%%%%%%
\subsection{Experiment 1: Concentric Spheres} \label{sec:sphere}
%%%%%%%%%%%%%%%%%%%%%%%%%%%%%%%%%%%%%%%%%%%%%%%%%%%%%%%%%%%%%%%%%%%%%%%%%%%%%%%%%%%%%%%%%%%%%%%%%%%%%%%%%%%%%%%%%%%%%%%%

The task in this experiment is binary classification on a dataset comprised of samples taken from two concentric $d$-dimensional spheres, respectively with radii 1 and 1.3 and carrying labels 0 and 1. While this is a small dataset which allows for rapid training and comparison, the non-linear nature of the decision boundary and the homology of the dataset make it a non-trivial problem. To simulate the presence of erroneous data, we flip the labels of a certain percentage of the samples as indicated below.

\begin{figure*}[ht]
    \centering
    % \fbox{
    \includegraphics[clip, trim=6.1cm 2.8cm 5.8cm 2.5cm,width=0.95\textwidth]{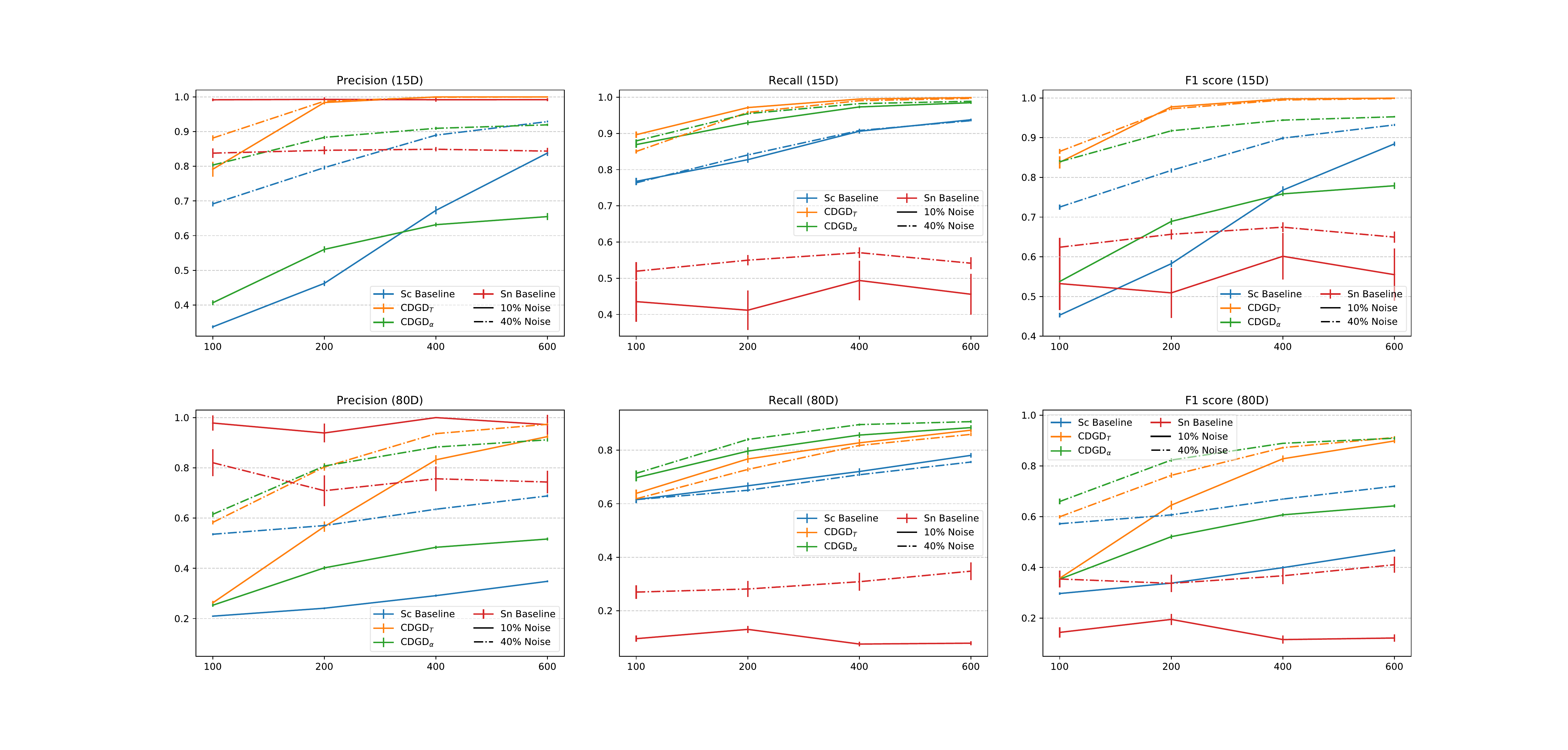}
    % }
    \caption{Precision, Recall and F1 score for 15 dimensional (top) and 80 dimensional spheres (bottom). In these plot, the line color indicates the method and the line style indicates the percentage of samples with flipped labels.}
    \label{fig:sphere_results}
\end{figure*}

\paragraph{Setup} For this experiment, we use a multi-layer perceptron (MLP) with one hidden layer of 1,000 rectified linear units (ReLU, \cite{glorot2011deep}). We use binary cross entropy loss which acts on the output of the network after a sigmoid function. No explicit regulator is used in this experiment, however, in all relevant experiments the clean dataset $\cl$ is used for early stopping. In this experiment, we do not use gradient truncation. During training of this experiment we use full batch vanilla gradient descent on the network parameters $\theta$ in the inner loop (i.e. without momentum etc.). In the outer loop, we use Adam to optimize the inclusion parameters $\alpha$. IThe inner loop uses a fixed learning rate of 1.4 throughout training but we use a learning rate scheduler which starts at 0.2 and reduces the learning rate by a factor of 10 if the loss has not decreased in the last 15 steps. The inner optimization loop runs for 600 epochs of the data which translates to 600 $\theta$ optimization steps and the outer loop performs 250 $\alpha$ optimization steps. For the baseline methods, we use Adam optimizer for 200 epochs learning rate scheduler which starts at 0.01 and reduces the learning rate by a factor of 10 if the loss has not decreased in the last 15 steps.

We arrived at these hyperparameter values by fixing the learning rate schedule, scanning over a range of learning rates and training until convergence and picking the hyperparameters with the best performance for each method. We then repeat the experiment with these hyperparameters 20 times for each method and report the mean and standard error of the results. In general we find very little variability between different runs which leads to a small standard error in this experiment.

\paragraph{Results} We first provide a visualization of our method and its main features by applying our methods to datasets defined in two dimensions (i.e. $d=2$). Fig.~\ref{fig:sphere_demo} shows the data and the cleanup results (i.e. $\no$ with samples determined to be erroneous removed) for a typical configuration with the noisy set $\no$ comprised of 100 samples and 40\% noise (top left). The clean set $\cl$ includes 20 correctly labeled samples (top right). 

Let us consider the results of each algorithm one by one. The $\no$ baseline (trained on $\no$ with $\cl$ as validation set for early stopping), determines the mislabeled data as the samples that are in the minority in any local neighborhood (middle left). As expected, this method performs poorly in areas where the mislabeled data are more numerous. The $\cl$ baseline (trained only on $\cl$) does slightly better in this example, but as seen in the figure, can misclassify in places where clean data is sparse (middle right). 

Our first method \thr{\alpha} (bottom left) makes a few mistakes, despite outperforming the baselines. If we consider the locations where \thr{\alpha} includes mislabeled data, for example the orange dot on the left side of the inner circle, these are again areas where the clean dataset $\cl$ is sparse.

%Note that this method is not equivalent to an ensemble of the two baselines. Although there are regions on the manifold where both baselines are incorrect (e.g. the upper right part of the inner circle), \thr{\alpha} does not make the same mistakes.

Finally, in this case our second method \fin, performs the best without making any mistakes (bottom right). This is despite the fact that it uses a classifer trained on the output of the \thr{\alpha} method which include two mislabeled datapoints.\footnote{
    More precisely \fin is trained on $\no$ with the $\alpha$ coefficients taken into account, however, even though $\alpha$ are continuous parameters between 0 and 1, after training they are often predominantly saturated toward either 0 or 1. In this case, of the 100 $\alpha$'s, only 4 take values not equal to 0 or 1 up to numerical precision. We can therefore think of \fin as being effectively trained on the output of \thr{\alpha}.
} 
 However, because these mislabeled points are now in the minority, they do not adversely affect the final outcome of \fin. We can therefore think of \fin as a smoothed out version of \thr{\alpha}.

For the rest of this experiment, we take the noisy dataset $\no$ to be 1,000 randomly generated samples.  We perform our experiment on a number of different configurations given by percentages of noisy samples in $\no$, dimensions of the sphere, and the number of samples in the clean dataset $\cl$. In each configuration we train 20 models and report the mean and standard error of the precision, recall and F1 scores of correctly identifying the flipped training samples.

\begin{figure*}[ht]
    \centering
    % \fbox{
    \includegraphics[clip, trim=0.4cm 0.5cm 0.5cm 0.0cm,width=0.9\textwidth]{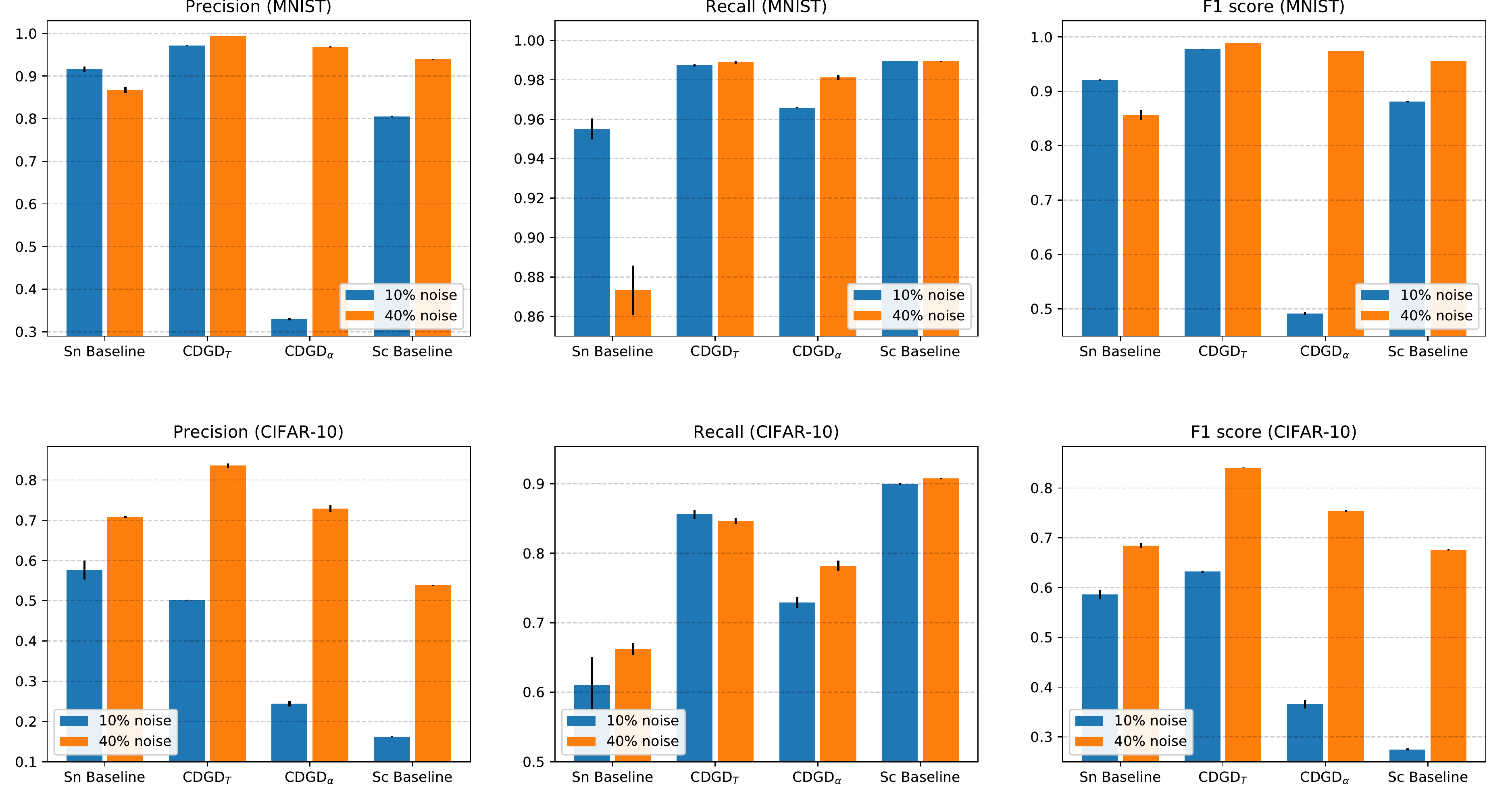}    
    % }
    \caption{Precision, recall and F1 score for finding the flipped samples in the MNIST (top) and CIFAR-10 (bottom) datasets.}
    \label{fig:cifar_mnist_results}
\end{figure*}

Fig. \ref{fig:sphere_results} gives the results for precision, recall and the F1 score for 15 and 80 dimensional sphere datasets with 10 and 40 percent noise levels in $\no$. We see that for the experiments with 10\% noise (solid lines), \fin far outperforms the other methods. For the experiments with 40$\%$ noise (dashed lines), on average, \thr{\alpha} performs the best followed by \fin and $\cl$ baseline trailing behind. In all of these experiments, the $\no$ baseline performs very poorly. This is due to the fact that when training on the entirety of the noisy dataset $\no$, the network tends to memorize the erroneous samples, leading to a very low recall rate and F1 score.

%%%%%%%%%%%%%%%%%%%%%%%%%%%%%%%%%%%%%%%%%%%%%%%%%%%%%%%%%%%%%%%%%%%%%%%%%%%%%%%%%%%%%%%%%%%%%%%%%%%%%%%%%%%%%%%%%%%%%%%%
\subsection{Experiment 2: MNIST and CIFAR-10} \label{sec:mnist_cifar}
%%%%%%%%%%%%%%%%%%%%%%%%%%%%%%%%%%%%%%%%%%%%%%%%%%%%%%%%%%%%%%%%%%%%%%%%%%%%%%%%%%%%%%%%%%%%%%%%%%%%%%%%%%%%%%%%%%%%%%%%

In this experiment we employ two popular machine learning benchmark datasets MNIST and CIFAR-10~\cite{krizhevsky2009learning} to test our methodology. We take the noisy dataset $\no$ to be the 50,000 training samples and flip 10 or 40 percent of the labels similar to the previous experiment.\footnote{
    Specifically, we take $y \to 9 - y$ for the flipped labels.
} 
We shuffle the test set once and take the first 5,000 samples to be the validation set $\cl$. We repeat the experiment in each configuration 4 times and report the mean and standard error of the result. 

\paragraph{Setup} Compared to the concentric sphere dataset, it is considerably more intensive computationally and memory wise to analyze the MNIST and especially the CIFAR-10 dataset. This is both because the input dimensions are higher and also in the case of CIFAR-10 we need a larger model. For these datasets we use a relatively small fully convolutional network comprised of 6 convolutional layers, and also resort to aggressive truncation which cuts off the gradient history of the $\alpha$ parameters after each epoch. 
With these considerations and also using half precision computation, we use the largest pyramid shape convolutional network that can be still fit on a GPU with 16GB of memory (Tab.~\ref{tab:models}). In this model, the first 5 convolutional layers are followed by a dropout layer with $p=0.2$. Apart from this no other regularization scheme is used. In this case, for the inner loop we use batch size 80 and train for 130 epochs with learning rate 0.12 and a learning rate schedule that reduces the learning rate by a factor of 10 if there is no decrease in the validation loss for 30 epochs. The outer loop learning rate is 0.05 with an Adam optimizer and 150 optimization steps and a scheduler which reduces the learning rate by 10 after a pleateau in the loss of 20 optimization steps. The baseline methods similarly use batch size of 80 but with an Adam optimizer and learning rate 0.12.

Similar to the sphere experiment, these hyperparameters were determined by fixing the learning rate schedule and scanning over a range of learning rates from 10 to 0.0001 by multiplicative increments of 3 and training until convergence and picking the hyperparameters with the best performance for each method. However, in this case we repeat the experiment in each configuration only 4 times because of the computation cost. We again find fairly small variability between different runs of the same configuration with the exception of the $\no$ baseline which can have large variability because of the tendency to memorize the noisy labels.

\begin{table}[h]
              \scalebox{1}{
              \begin{tabular}{cccccc}
                    \hline\hline
                    Layer               &  Channels    &    Kernels        &   Strides     &   Padding          \\
                    \hline
                    Convolution             &   48      &   $5\times5$    &   2       &      1                  \\       
                    Convolution             &   64      &   $3\times3$    &   2       &      1                  \\
                    Convolution             &   96      &   $3\times3$    &   1       &      0                  \\
                    Convolution             &   128     &   $3\times3$    &   2       &      1                  \\
                    Convolution             &   168     &   $3\times3$    &   1       &      0                  \\
                    Convolution            &   10       &   $3\times3$    &   2       &      1                  \\
                    AvgPool                                                                                     \\   
                    \hline
              \end{tabular}}
% 		}
	\vspace{8pt}
	\caption{
% 	\small 
	Network structure for MNIST and CIFAR-10.}
% 	\vspace{-20pt}
	\label{tab:models}
\end{table}

\paragraph{Results} The results for the MNIST and CIFAR-10 benchmarks are given in Fig. \ref{fig:cifar_mnist_results}. We see that \fin performs the best in all scenarios, with a few exceptions where it matches the best performance within the margin of error. However, compared to the sphere results, we see a few major differences. First, the $\no$ baseline performs significantly better in the MNIST and CIFAR-10 experiments. We suspect this is because the size and structure of the models, as well as the number of training samples are such that they do not easily allow the network to memorize the flipped labels. We expect that if we used a significantly larger network akin to the state-of-the-art models, the $\no$ baseline method would suffer the same fate as before.\footnote{
    In fact there are a number of works in the literature that show that large neural nets are expressive enough to memorize large datasets with completely randomized labels~\cite{memorize}, making them unsuitable for the purposes of sample verification using the $\no$ baseline.
}
Second, we see that the $\cl$ baseline model performs worse in the CIFAR-10 experiment compared to the cases of MNIST and sphere. This is most likely because the size of the $\cl$ is not large enough to sufficiently train a classification model on CIFAR-10. We also note that the method \thr{\alpha} performs poorly in low noise configurations on  both MNIST and CIFAR-10 datasets.

The most challenging test, CIFAR-10 with 40\% noise demonstrates the power of our method. Here, \fin provides a relative improvement of about 25\% over the F1 score of the best baseline method. 

\section{Conclusion}
\label{sec:conclusion}

In this paper we introduced a new gradient descent based method for finding mislabeled or erroneous samples and discussed its place along the typical machine learning project work flow. The algorithm works by determining which of the samples in a noisy training set to keep and which to discard in order to optimize the performance of a small clean dataset. This procedure is made possible by generalizing the inclusion parameters to continuous variables and determining them via gradient descent. 

We conducted an empirical study of the properties and performance of this new method under synthetic and real world datasets.  In general, we find that compared to the $\no$ and $\cl$ baselines, two commonly used methods which simply train a neural network on the entirety of the noisy or clean datasets respectively, our methods provide considerably better results. They  are, however, considerably more resource intensive. We discussed various techniques including gradient truncation and checkpointing schemes which  make it possible to work with larger models and datasets. Finally, we point out that it would be interesting to understand this approach more theoretically as an optimization on $\cl$ which is constrained by $\no$.

\begin{acknowledgments}
We would like to thank Kyle Cranmer for interesting discussions and input, Adam Paszke and Soumith Chintala for their help with PyTorch. SG is supported by the James Arthur Postdoctoral Fellowship.
\end{acknowledgments}

\appendix

% \section{Sphere dataset detailed results}\label{app:sphere_results}

% Any more results go here.

% \begin{figure}[ht]
%     \centering
%     \includegraphics[clip, trim=4cm 1.5cm 3.5cm 1.5cm,width=\textwidth]{sphere_results_precision.pdf}    \caption{Precision for 15 (left) and 80 (right) dimensional spheres. In this plot, the line color indicates the method and the line style indicates the percentage of samples with flipped labels.}
%     \label{fig:sphere_precision}
% \end{figure}
% \begin{figure}[ht]
%     \centering
%     \includegraphics[clip, trim=4cm 1.5cm 3.5cm 1.5cm,width=\textwidth]{sphere_results_recall.pdf}    \caption{Recall for 15 (left) and 80 (right) dimensional spheres. In this plot, the line color indicates the method and the line style indicates the percentage of samples with flipped labels.}
%     \label{fig:sphere_recall}
% \end{figure}

% \input{sphere_appendix}

\section{Threshold Dependence of \thr{\alpha}} \label{app:pnr}

\begin{figure}[ht]
    \centering
    \includegraphics[clip, trim=0.3cm 0.1cm 0.3cm 0.1cm,width=0.35\textwidth]{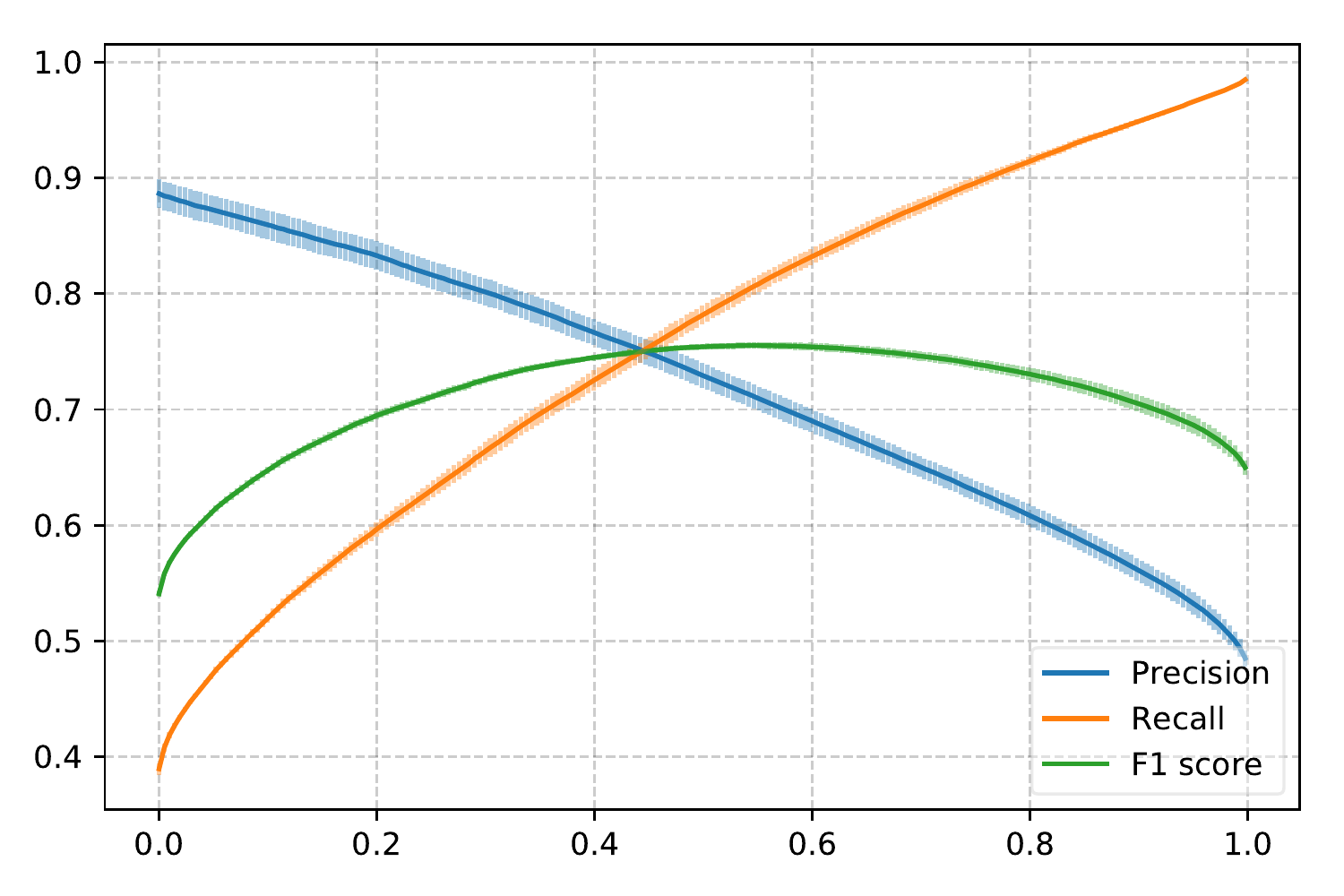} 
    \caption{Precision, recall and F1 score curves as a function of the threshold parameter. Shaded region denotes the standard error.}
    \label{fig:cifar_pnr}
\end{figure}

During the experiments in the main body of this paper we took the threshold parameter in the \thr{\alpha} method to be $0.5$. Here, we provide the precision and recall curves for the selection of the threshold parameter. Fig.~\ref{fig:cifar_pnr} shows the typical shape of these results, in this case for CIFAR-10 with 40\% noise. In this example, the F1 score curve achieves its maximum at threshold value $0.54$. 

With respect to the determination of mislabeled data points, strictly speaking, only samples $x_i$ whose $\alpha_i$ is exactly equal to zero, are completely excluded from training. Because of this, we might be tempted to define the noisy samples as those which have $\alpha_i = 0$, i.e. set the threshold $0$.  However, in Fig.~\ref{fig:cifar_pnr}, we see that this value of the threshold in fact performs poorly.

\bibliographystyle{ACM-Reference-Format}
\bibliography{biblio}
\onecolumngrid
% \clearpage

\end{document}